\documentclass[conference]{IEEEtran}
\usepackage{times}

\usepackage[numbers]{natbib}
\usepackage{multicol}
\usepackage{amsmath}
\usepackage{amssymb}
\usepackage{amstext}
\usepackage{array}
\usepackage{graphicx}
\usepackage[caption=false]{subfig}
\usepackage[bookmarks=true]{hyperref}
\usepackage{microtype}
\usepackage{adjustbox}
\usepackage{booktabs}
\usepackage{placeins}
\usepackage{balance}

\usepackage{color}
\usepackage{colortbl}

\definecolor{yellow}{rgb}{1,1,0.8}
\definecolor{blue}{rgb}{0.8,0.9,1.0}
\definecolor{green}{rgb}{0.7,0.95,0.65}
\definecolor{gray}{rgb}{0.9,0.9,0.9}

\usepackage{bm}

\def\eqref#1{equation~\ref{#1}}
\def\1{\bm{1}}

\def\vtheta{{\bm{\theta}}}

\def\vk{{\bm{k}}}

\def\vq{{\bm{q}}}

\def\vv{{\bm{v}}}

\def\vx{{\bm{x}}}
\def\vy{{\bm{y}}}
\def\vz{{\bm{z}}}

\DeclareMathAlphabet{\mathsfit}{\encodingdefault}{\sfdefault}{m}{sl}
\SetMathAlphabet{\mathsfit}{bold}{\encodingdefault}{\sfdefault}{bx}{n}

\def\sR{{\mathbb{R}}}

\newcommand\blfootnote[1]{%
  \begingroup
  \renewcommand\thefootnote{}\footnote{#1}%
  \addtocounter{footnote}{-1}%
  \endgroup
}

\begin{document}

% paper title
\title{Towards Differentiable Resampling}

\author{\authorblockN{Michael Zhu$^{\,1}$}
\authorblockA{Stanford\\
mhzhu@stanford.edu}
\and
\authorblockN{Kevin Murphy}
\authorblockA{Google AI\\
kpmurphy@google.com}
\and
\authorblockN{Rico Jonschkowski}
\authorblockA{Robotics at Google\\
rjon@google.com}}

\maketitle

%%%%%%%%%%%%%%%%%%%%%%%%%%%%%%%%%%%%%%%%%%%%%%%%%%%%%%%%%%%%%%%%%%%%%%%%%%%%%%%%
\begin{abstract} 
Resampling is a key component of sample-based recursive state estimation in particle filters. Recent work explores \emph{differentiable} particle filters for end-to-end learning~\cite{DBLP:conf/rss/JonschkowskiRB18,pmlr-v87-karkus18a}. However,  resampling remains a challenge in these works, as it is inherently non-differentiable. We address this challenge by replacing traditional resampling with a learned neural network resampler. We present a novel network architecture, the \emph{particle transformer}, and train it for particle resampling using a likelihood-based loss function over sets of particles. Incorporated into a differentiable particle filter, our model can be end-to-end optimized jointly with the other particle filter components via gradient descent. Our results show that our learned resampler outperforms traditional resampling techniques on synthetic data and in a simulated robot localization task.
\end{abstract}

\IEEEpeerreviewmaketitle

%%%%%%%%%%%%%%%%%%%%%%%%%%%%%%%%%%%%%%%%%%%%%%%%%%%%%%%%%%%%%%%%%%%%%%%%%%%%%%%%
\section{Introduction}

State estimation, estimating the posterior distribution of the current state given past observations and actions, remains a key problem in robotics. This problem is often addressed with particle filters~\cite{gordon1993novel,liu1998sequential}, which approximate the posterior by a set of weighted samples (particles) that are updated at each time step. Particles are moved according to the predictions of a motion model for the current action, and their weights are updated based on a measurement model that estimates the likelihood of the current observation for each particle. Finally, particles are resampled according to their weights to focus the approximation of the posterior to regions of high probability. Traditionally, particle filters required manually defined or individually learned motion and measurement models.\blfootnote{$^{1\,}$This work was done while at Robotics at Google.}

Recent work on differentiable particle filters (DPFs)~\cite{DBLP:conf/rss/JonschkowskiRB18,pmlr-v87-karkus18a} replaces traditionally handcrafted models by neural networks and implements the particle filter algorithm in a way that allows computing the gradient by differentiation. As a result, DPFs can optimize models for end-to-end state estimation by gradient descent. DPFs can also be viewed as recurrent neural networks (RNNs) that leverage the problem structure in state estimation to improve data efficiency and generalization~\cite{DBLP:conf/rss/JonschkowskiRB18}.

Albeit promising results, DPFs are limited by the inherently non-differentiable resampling step. That step is necessary because repeatedly updating the positions and weights of particles eventually leads to a degeneracy problem where most particles have close to zero weight, and only a few of the particles are useful for approximating the posterior. To mitigate this problem, particles are randomly resampled with replacement according to their weights to generate a uniformly-weighted particle set that represents the same posterior, which is then used as a prior for the next time step. The resampled particles are differentiable with respect to the poses of the input particles from which they were copied (their ancestors), but they are not differentiable with respect to the input weights. For this reason, traditional resampling does not allow backpropagating the gradient in differentiable particle filters across multiple time steps. Previous work avoids this issue by truncating backpropagation after a single step~\cite{DBLP:conf/rss/JonschkowskiRB18}, by training without any resampling when possible (e.g. for tracking from a known state)~\cite{pmlr-v87-karkus18a}, or by soft resampling~\cite{pmlr-v87-karkus18a}, which trades resampling quality for a biased gradient estimate (see Section~\ref{sec:background_resampling}). As none of these measures solve the differentiability problem, we investigate replacing traditional resampling with a learned neural network resampler.

In addition to enabling differentiability, our approach also raises an important question: \emph{Is traditional resampling optimal for state estimation}, especially when models are learned end-to-end? For example, it is known in practice~\cite[p.~118]{thrun2005probabilistic} and has been found empirically through end-to-end learning~\cite{DBLP:conf/rss/JonschkowskiRB18} that particle filters perform best if motion and measurement models \emph{overestimate} the noise of the system. One reason why this might help is that it increases robustness to unmodeled effects. But increased noise might also help by compensating for approximation errors from particle resampling. 

Since resampling produces exact copies of highly weighted particles, the motion model \emph{must be noisy} to spread these particles out to enable the measurement update to assign different weights to them, \emph{even if the true motion is deterministic}. One could imagine an alternative way of resampling that does not duplicate particles but instead interpolates between them to spread particles out from the start, which might reduce the required noise in the motion model. Although importance resampling is optimal under certain assumptions~\cite{thrun2005probabilistic}, it is an intriguing question if there is a form of resampling that works better empirically (where some assumptions might not hold).

Our contributions are: (1) We replace existing resampling mechanisms with a neural network to enable differentiability and to study whether learned models can outperform traditional resampling. (2) We present the \emph{particle transformer}, a new network architecture tailored to particle resampling. (3) We present two ways of training this model to perform resampling using (3a) a likelihood-based loss function between sets of particles and (3b) end-to-end training of the resampling model as part of the differentiable particle filter. Our experiments show that the learned model outperforms existing resampling methods on synthetic data and in a simulated localization task.

%%%%%%%%%%%%%%%%%%%%%%%%%%%%%%%%%%%%%%%%%%%%%%%%%%%%%%%%%%%%%%%%%%%%%%%%%%%%%%%%
\section{Related Work}
 Differentiable Particle Filters~\cite{DBLP:conf/rss/JonschkowskiRB18} and Particle Filter Networks~\cite{pmlr-v87-karkus18a} combine the particle filter algorithm with learnable motion and measurement models and end-to-end training by gradient descent. Particle Filter RNNs~\cite{ma2019particle} generalize this approach beyond classical state estimation problems through an architecture that is closer to vanilla RNNs but maintains a particle-based latent distribution instead of a latent vector. As discussed in the introduction and in Section~\ref{sec:background_resampling}, differentiable resampling remains a key challenge, which our work addresses by proposing to learn resampling in a neural network.

Neural networks can model point sets  by first applying point-wise transformations and then aggregating the resulting point features by pooling~\cite{qi2017pointnet,zaheer2017deep}. Set transformers~\cite{lee2019set} build on these ideas and use a permutation-invariant encoder-decoder architecture that leverages multi-head attention~\cite{vaswani2017attention}. Our work extends set transformers for resampling by introducing weighted attention and scale-equivariance.

%%%%%%%%%%%%%%%%%%%%%%%%%%%%%%%%%%%%%%%%%%%%%%%%%%%%%%%%%%%%%%%%%%%%%%%%%%%%%%%%
\section{Background}
\label{sec:background}

\subsection{Resampling in Particle Filters}
\label{sec:background_resampling}

Particle filters recursively estimate the posterior probability over states, given a sequence of observations and actions. They represent this posterior by a set of weighted samples $\{\vx^{(i)},w^{(i)}\}_{i=1}^{n}$, also called particles, where $\vx^{(i)}\in\sR^d$ represents the $d$-dimensional sample position and $w^{(i)}\in\sR$ represents the normalized particle weight for particle $i$ such that $\sum_{i=1}^n w^{(i)} = 1$. Repeated motion and measurement updates without resampling would spread out particles and reduce most weights until only a few particles have non-negligible weight, at which point the particle set is degenerated and cannot properly estimate the posterior anymore. Resampling addresses this problem by focusing particles in high-probability regions~\cite{thrun2005probabilistic}. Traditional particle filters use multinomial or systematic resampling. Soft resampling is a differentiable version of these ideas~\cite{pmlr-v87-karkus18a}. All of these methods perform resampling by removing or duplicating existing particles.

\emph{Multinomial resampling} draws particles with replacement according to the discrete distribution represented by the particle weights. \emph{Systematic resampling} (also \emph{stochastic universal sampling}) turns the particle weights into a cumulative distribution, uniformly samples an offset $\delta\in[0,\frac{1}{n})$, and draws particles whenever their component of the cumulative distribution includes $\delta + \frac{k}{n}$ for integers $k\in[0, n-1]$. By introducing a dependency between the generated samples, this procedure increases stability of the particle set as every particle with normalized weight $>\frac{c}{n}$ is guaranteed to be sampled at least $c$ times. \emph{Soft resampling}~\cite{pmlr-v87-karkus18a} is a differentiable approximation that performs multinomial resampling on a mixture of the weight distribution and a uniform distribution, and then mixes the weights of the sampled ancestors into the uniform weights of the resampled particles. This makes the weight of the resampled particle differentiable with respect to the weight of its ancestor. Unfortunately, soft resamplng biases both the resampling process and the estimated gradients, and it only provides gradients for input particles that were sampled.

\subsection{Transformer / Set Transformer Models}

The \emph{transformer}~\cite{vaswani2017attention} model is a neural network architecture centered around an \emph{attention} function that maps a query and a set of key-value pairs to a weighted sum of values, where each weight is computed as the similarity between query and key (further described in Section~\ref{sec:particle_transformer}). The transformer model generalizes this idea to \emph{multi-head attention} by linearly projecting each query, key, and value in $h$ different ways, applying the attention function to all $h$ query-key-value triplets, concatenating the results, and linearly projecting them again. Transformers interleave multi-head attention with point-wise feed-forward layers, residual connections, and layer normalization into an encoder-decoder architecture. When applied to natural language processing, input words are represented with learned embeddings and encodings of their position.

The \emph{set transformer}~\cite{lee2019set} model extends the transformer to leverage attention for learning interactions in point sets. To achieve permutation-invariance on the input set, this model omits positional encodings. On the decoder side, the set transformer proposes \emph{pooling by multi-head attention} (PMA), which uses $k$ learnable vectors as seeds for queries in multi-head attention. The decoder aggregates the encoder outputs using PMA with $k$ seed vectors, multi-head self-attention, and point-wise feed-forward layers to generate $k$ outputs.

%%%%%%%%%%%%%%%%%%%%%%%%%%%%%%%%%%%%%%%%%%%%%%%%%%%%%%%%%%%%%%%%%%%%%%%%%%%%%%%%
\section{Particle Transformers for\\Differentiable Resampling}
\subsection{Particle Transformer}
\label{sec:particle_transformer}

We propose the \emph{particle transformer}, a neural network architecture for particle resampling that is both permutation-invariant and scale-equivariant, and incorporates particle weights in the attention mechanism. Our model builds on the set transformer~\cite{lee2019set} described above. However, instead of mapping a set of $n$ points to $k$ output vectors, our model maps a set of $n$ weighted points to a new set of $n$ points and leverages additional problem structure in particle resampling by incorporating scale-equivariance and weighted attention.

To appropriately model particle weights, we introduce a \emph{weighted attention} mechanism that generalizes dot-product attention. Suppose we have a query vector $\vq\in\mathbb{R}^{d_{k}}$ and $m$ pairs of keys $\vk^{(i)}\in\mathbb{R}^{d_{k}}$ and values $\vv^{(i)}\in\mathbb{R}^{d_{v}}$ for $i\in\left[1,m\right]$. The scaled dot-product attention function computes $\frac{\sum_{i=1}^{m}\exp(\vq\cdot \vk^{(i)}/\sqrt{d_{k}})\vv^{(i)}}{\sum_{j=1}^{m}\exp(\vq\cdot \vk^{(j)}/\sqrt{d_{k}})\hphantom{\vv^{(i)}}}$. We generalize this function to incorporate weights $w^{(i)}\in\mathbb{R}$ for each of the $m$ key-value pairs $\vk^{(i)},\vv^{(i)}$ to $\frac{\sum_{i=1}^{m}w^{(i)}\exp(\vq\cdot \vk^{(i)}/\sqrt{d_{k}})\vv^{(i)}}{\sum_{j=1}^{m}w^{(j)}\exp(\vq\cdot \vk^{(j)}/\sqrt{d_{k}})\hphantom{\vv^{(i)}}}$. We extend this idea to \emph{weighted multi-head attention} by using multiple key-value-query triplets as in multi-head attention~\cite{vaswani2017attention}.

\begin{figure}
\centering
\includegraphics[width=0.7\columnwidth]{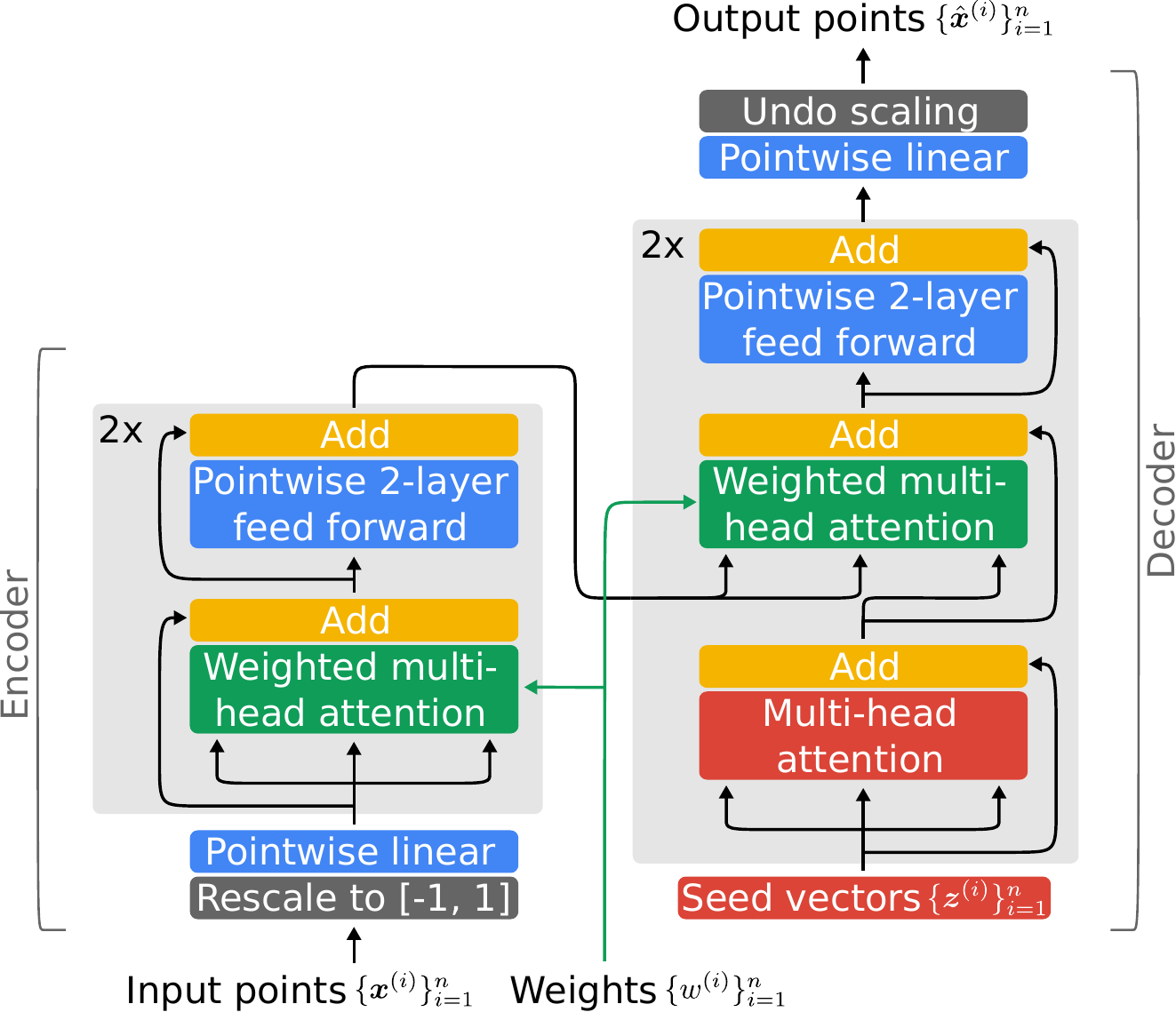}
\caption{Particle transformer architecture. Multi-head attention modules take three inputs: key, value, and query (left to right). Seed vectors are learned.}
\label{fig:model}
\vspace{-0.3cm}
\end{figure}

Our model uses weighted and vanilla multi-head attention as shown in Figure~\ref{fig:model}. The encoder takes as input a set of $n$ weighted vectors $\{\vx^{(i)}, w^{(i)}\}_{i=1}^n$, scales each input dimension based on their minima and maxima to $[-1, 1]$, then linearly transforms each vector into a latent representation. It then twice applies weighted multi-head self-attention and pointwise feed-forward layers. Up to this point, each encoded vector is still associated with its original weight. The decoder takes these encoded particles as input, together with $n$ learned vectors $\vz^{(i)}$ that act as \textquotedblleft seed vectors\textquotedblright{}~\cite{lee2019set}, to generate $n$ new particles. The decoder applies the following twice: multi-head self-attention, weighted multi-head attention -- where encoded particles are used to compute keys and values and the decoder latents are used to compute queries -- and pointwise feed-forward layers. The resulting $n$ vectors are linearly transformed to match the input dimensionality and rescaled per dimension to their original minima and maxima. Throughout the model we use a latent dimension of 256 and multi-head attention with 8 heads. 

\subsection{Loss Function for Individual Training}

After having introduced the particle transformer, which maps $n$ weighted particles to $n$ equally-weighted particles, we will now define the loss function for training this model for resampling. The idea is to maximize the likelihood of a set of target particles conditioned on the resampled particles. To convert a set of particles into a probability density function, we apply kernel density estimation and compute a mixture of Gaussians (with one Gaussian per particle)~\cite{DBLP:conf/rss/JonschkowskiRB18}. Let $q_{\text{r}_{\vtheta}(\{\vx^{(i)},w^{(i)}\}_{i=1}^{n})}$ be the Gaussian mixture distribution of the $n$ particles resampled by the model with parameters $\vtheta$ from input particles $\{\vx^{(i)},w^{(i)}\}_{i=1}^{n}$ and let $\{\vy^{(i)},v^{(i)}\}_{i=1}^{n}$ denote a set of target particles. The loss can be derived from the Kullback-Leibler (KL) divergence between the predicted and the target distributions (see Appendix~\ref{app:loss}). It sums the appropriately weighted $q$'s at the positions of the $n$ target particles
\begin{equation}
L(\theta)=-\sum_{i=1}^{n}\text{\ensuremath{\frac{v^{(i)}}{\sum_{j=1}^{n}v^{(j)}}}}\log q_{\text{r}_{\vtheta}(\{\vx^{(i)},w^{(i)}\}_{i=1}^{n})}(\vy^{(i)}).
\label{eq:loss}
\end{equation}
As target particle sets, we use either (i) the input particles $\{\vx^{(i)},w^{(i)}\}_{i=1}^{n}$ or (ii) the output of a traditional resampler on that input. Note that (i) only works because the particle transformer is constrained to output uniformly weighted particles, which prevents it from learning the identity mapping.

\subsection{Individual and End-to-End Training with Particle Filters}
\label{sec:method-e2e-training}

To study the interaction of differentiable resampling with end-to-end training in particle filters, we integrate our particle transformer with prior work on differentiable particle filters (DPFs)~\cite{DBLP:conf/rss/JonschkowskiRB18}. We follow their recipe of training all particle filter components first individually and then end-to-end. Since individual training of the resampler requires examples of sets of input and target particles, we collect such a dataset by running the filter with individually trained motion and measurement models and a standard resampler on the problem of interest. Collecting these examples in the particle filter ensures that the training distribution of particle sets is similar to the distribution encountered during filtering. The exact steps for individual and end-to-end training are:
\begin{enumerate}
\item Individually train the motion and measurement models.
\item Run the particle filter from 1) with a standard resampler. Collect the resampler's input and output particle sets.
\item Using the training examples from step 2), individually train the particle transformer for resampling.
\item Train the DPF end-to-end to jointly optimize the motion model, the measurement model, and the resampler.
\end{enumerate}

%%%%%%%%%%%%%%%%%%%%%%%%%%%%%%%%%%%%%%%%%%%%%%%%%%%%%%%%%%%%%%%%%%%%%%%%%%%%%%%%
\section{Experiments}
\begin{figure*}
\includegraphics[width=1\linewidth]{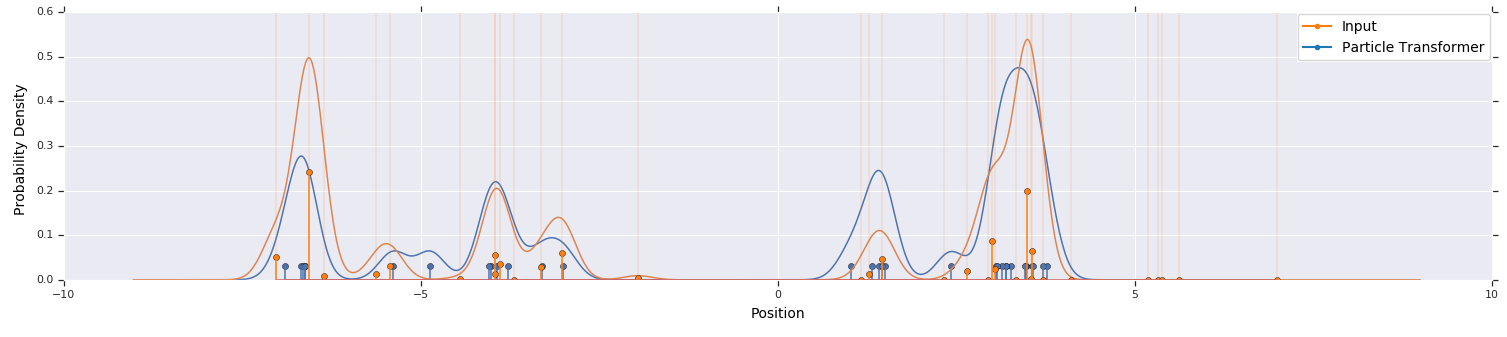}
\vspace{-0.9cm}
\caption{Qualitative results on synthetic data (one of five dimensions shown). Circles denote particle positions and weights. Lines show kernel density estimates.}
\label{fig:synthetic_qualitative}
\vspace{-0.3cm}
\end{figure*}

\begin{figure}
\includegraphics[width=0.95\columnwidth]{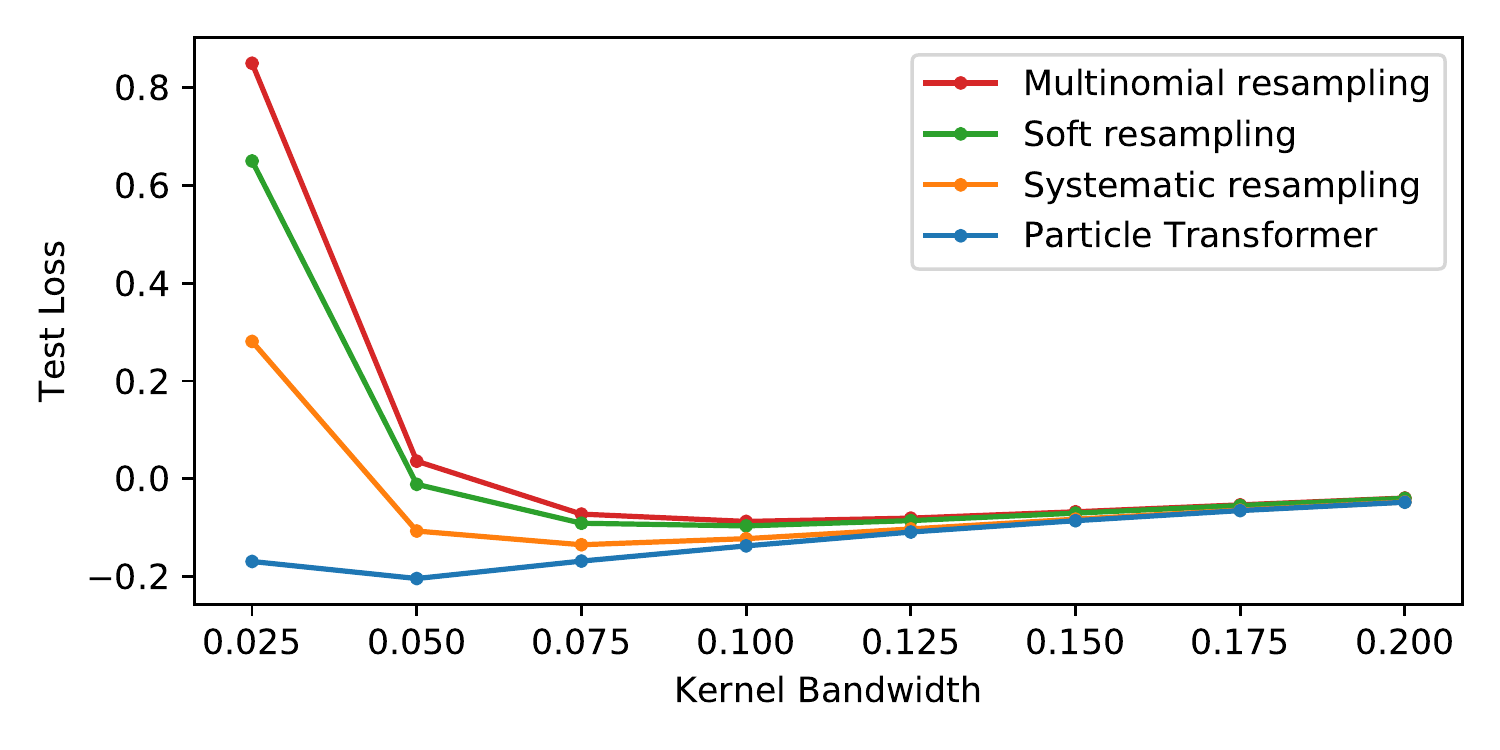}
\vspace{-0.5cm}
\caption{Comparison of different resampling methods on synthetic data.}
\label{fig:synthetic_loss}
\end{figure}

\subsection{Synthetic Data}

We first compare the particle transformer to other resampling techniques on a synthetic dataset of particle sets. We create this dataset by randomly generating pairs of probability distributions, one to generate samples and one to assign weights to those samples. We define these distributions as mixtures of three Gaussians in five-dimensional space, and randomly draw means from {[}-5, 5{]} and standard deviations from {[}1, 3{]} for each dimension. We sample mixture probabilities from {[}0.2, 0.4{]} for the first two Gaussians, which determine the probability of the last Gaussian such that all three sum to one. We use 32 particles drawn from these distributions as both input and target sets, and generate and generate 60k particle sets -- 50k for training, 10k for evaluation.

Figure~\ref{fig:synthetic_qualitative} shows qualitative results of the particle transformer in a one-dimensional slice of the five-dimensional distribution (inputs are orange, outputs are blue). Output particles are constrained to have equal weights to force them to focus on high-probability regions. Although this constraint hinders approximating the input distribution, the resampled particles capture all modes. Note that unlike traditional resampling techniques that replicate or remove particles, our learned resampler places particles where there were no input particles, which increases particle diversity (e.g. in the rightmost mode).

Figure~\ref{fig:synthetic_loss} compares the particle transformer to other resampling techniques -- multinomial resampling, systematic resampling, and soft resampling -- in terms of the loss defined in Equation~\ref{eq:loss} on the evaluation set. We compute this loss for different values of the kernel bandwidth, which controls the width of the Gaussians centered at each of the particles for estimating $q$. The results show that the particle transformer achieves the lowest loss for all kernel bandwidths, especially for small bandwidths, where it achieves a loss of about -0.2 compared to 0.3 to 0.8 for the other methods. Multinomial resampling, which is most aggressive at removing particles, fares poorly in this setting. Soft resampling that mixes weights with a uniform distribution provides a small improvement. Systematic resampling, which keeps input particles above a certain weight, works better yet. However, our learned resampling strategy -- which is not constrained to replicate or remove particles -- substantially outperforms all other methods.

\subsection{Differentiable Resampling for Particle Filters}

To test our learned resampler in a particle filter, we consider the problem of robot localization in a simulated maze environment. As the robot moves through the maze, the goal is to estimate the robot's pose from a sequence of camera observations and odometry measurements. We use the dataset and the differentiable particle filter implementation from Jonschkowski et al.~\cite{DBLP:conf/rss/JonschkowskiRB18}, but use their particle proposer network only to initialize particles in the first time step. For all other steps, we resample all 100 particles with the particle transformer (or other methods, when we compare to those).

\begin{table}
\caption{Particle filter results for individual $\to$ end-to-end training}
\label{tab:main_table}
\vspace{-0.3cm}
\centering{}%
\adjustbox{width=\linewidth}{
    \begin{tabular}{lll}
    \toprule
    Resampling method & Error rate at last time step & MSE at last time step\\
    \midrule
    No resampling & 70.0$\pm$0.3\% $\to$ 58.0$\pm$0.3\%
    & 7.0$\pm$0.2 $\to$ 3.5$\pm$0.03\\
    Soft resampling & 11.4$\pm$0.4\% $\to$ 6.9$\pm$0.2\% & 4.9$\pm$0.2 $\to$ 5.7$\pm$0.2\\
    Soft resampling (with BPTT) & 11.4$\pm$0.4\% $\to$ 6.6$\pm$0.2\% & 4.9$\pm$0.2 $\to$ 5.5$\pm$0.2\\
    Systematic resampling & \phantom{0}{\bf7.0}$\pm$0.2\% $\to$ 4.9$\pm$0.1\% & 2.9$\pm$0.2 $\to$ 3.9$\pm$0.1\\
    Particle transformer (frozen) & 12.0$\pm$0.7\% $\to$ 4.7$\pm$0.2\% & {\bf2.6}$\pm$0.1 $\to$ 2.3$\pm$0.08 \\
    Particle transformer & 12.0$\pm$0.7\% $\to$ {\bf1.2}$\pm$0.2\% & {\bf2.6}$\pm$0.1 $\to$ {\bf0.38}$\pm$0.06\\
    \bottomrule
    \end{tabular}
}
\vspace{-0.3cm}
\end{table}

We train all particle filter components first individually and then optimize them end-to-end as described in Section~\ref{sec:method-e2e-training}. In this experiment we stop gradients across time steps such that there is no backpropagation through time (BPTT), save for soft resampling which we evaluate with and without BPTT. Systematic resampling method does not allow BPTT as it is not differentiable. Particle transformers are designed to provide gradients for BPTT, but are currently limited in this regard due to exploding gradients (see Appendix~\ref{app:bptt}).

Table \ref{tab:main_table} shows the test results for different resampling algorithms for two metrics, error rate and mean-squared error (MSE), evaluated after individual training, and then again after additional end-to-end training. All results are given with standard errors from five independent trials in Maze 1~\cite{DBLP:conf/rss/JonschkowskiRB18}. The resulting error rates show that resampling is crucial for this localization problem with error rate reductions from 70\% to 7\% after individual training and from 58\% to 1.2\% after end-to-end training. We also see that systematic resampling consistently outperforms soft resampling (7\% vs. 11.4\% and 4.9\% vs. 6.6\%), and we see only a minor benefit from the BPTT with soft resampling (6.6\% vs. 6.9\%), potentially due to biased gradients as described in Section~\ref{sec:background_resampling}. Systematic resampling also outperforms our learned resampler if all models (resampling, motion model, and measurement model) are trained individually (7\% vs. 12\%), apparently because the individual learning objectives for the motion and measurement models are more aligned with this traditional way of resampling. The results change, however, after end-to-end training. When we train the motion and measurement models end-to-end (but freeze the resampling model), our model achieves comparable error (4.7\% vs. 4.9\%) rates and lower mean squared errors (2.3 vs. 3.9). If we additionally train the resampling model end-to-end, the results improve substantially again. Compared to systematic resampling, our model reduces the error rate from 4.9\% to 1.2\% and the mean squared errors from 3.9 to 0.38 (qualitative results in Appendix~\ref{app:qualitative_results}).

%%%%%%%%%%%%%%%%%%%%%%%%%%%%%%%%%%%%%%%%%%%%%%%%%%%%%%%%%%%%%%%%%%%%%%%%%%%%%%%%
\section{Conclusion}
We have shown progress towards differentiable resampling in particle filters and tested whether learned models can provide better resampling than traditional approaches. To explore learning for the resampling problem, we introduced the particle transformer, a permutation-invariant and scale-equivariant model with weighted attention that can be optimized for end-to-end state estimation. Our results show that this learned model substantially outperforms traditional resampling methods on synthetic data and in a simulated localization task. Future work is needed to enable backpropagation through time in order to realize the full potential of differentiable resampling.

%%%%%%%%%%%%%%%%%%%%%%%%%%%%%%%%%%%%%%%%%%%%%%%%%%%%%%%%%%%%%%%%%%%%%%%%%%%%%%%%
\pagebreak
\nobalance
\appendix
\section{Appendix}
\FloatBarrier
\subsection{Derivation of loss function}
\label{app:loss}

We derive the loss function for one training example as follows. Let $p$ be the target distribution specified by the probability measure of the $n$ particle positions and their associated weights $\{\vy^{(i)},v^{(i)}\}_{i=1}^{n}$,
\[
p(\vy)=\sum_{i=1}^{n}\text{\ensuremath{\frac{v^{(i)}}{\sum_{j=1}^{n}v^{(j)}}}}\delta_{\vy^{(i)}}(\vy),
\]
where $\delta_{\vy^{(i)}}(\vy)$ denotes the Dirac delta function located at $\vy^{(i)}$, the position of particle $i$. 

Let $q_{\text{r}_{\vtheta}(\{\vx^{(i)},w^{(i)}\}_{i=1}^{n})}$ be the Gaussian kernel density estimate (mixture distribution) of the $n$ particles output by the particle transformer with parameters $\vtheta$ and input particles $\{\vx^{(i)},w^{(i)}\}_{i=1}^{n}$. We want to minimize the KL divergence between $p$ and $q_{\text{r}_{\vtheta}(\{\vx^{(i)},w^{(i)}\}_{i=1}^{n})}$, or equivalently the cross entropy
\[
\min_{\vtheta}\text{KL}(p\|q_{\text{r}_{\vtheta}(\{\vx^{(i)},w^{(i)}\}_{i=1}^{n})})=\min_{\vtheta}\text{H}(p\|q_{\text{r}_{\vtheta}(\{\vx^{(i)},w^{(i)}\}_{i=1}^{n})}).
\]
Hence, our loss function for one training example is
\begin{align*}
L(\vtheta) & =-\int p(\vx)\log q_{\text{r}_{\vtheta}(\{\vx^{(i)},w^{(i)}\}_{i=1}^{n})}(\vx)d\vx\\
 & =-\sum_{i=1}^{n}\text{\ensuremath{\frac{v^{(i)}}{\sum_{j=1}^{n}v^{(j)}}}}\log q_{\text{r}_{\vtheta}(\{\vx^{(i)},w^{(i)}\}_{i=1}^{n})}(\vy^{(i)}).
\end{align*}

\subsection{Towards Enabling Backpropagation Through Time}
\label{app:bptt}

Since the particle transformer is fully differentiable, it \emph{in principle} supports end-to-end learning with backpropagation through time (BPTT), i.e., we can unroll differentiable particle filters with particle transformer resampling across multiple time steps and have the gradients flow through the learned resampling mechanism to estimate how model parameters at time $t$ affect particle filter outputs at time $t+d$. This is exactly what motivated our work in the first place -- optimizing all particle filter components for their effect on all future time steps. Unfortunately, BPTT does not improve performance, which is why we stopped gradients from flowing across time steps when we evaluated end-to-end training of the particle transformer in the differentiable particle filter (shown in Table~\ref{tab:main_table}).

To better understand this issue, we performed experiments with and without freezing the resampler during end-to-end training, where we stopped gradients every $k$ time steps. I.e., we stop the gradients after resampling at step $t$ if $t$ mod $k$ $==$ $0$. The results for $k\in\{1,3,5\}$ in Table~\ref{tab:mod} show that if we freeze the resampler, performing this form of truncated BPTT with $k=3$ improves performance over stopping gradients at every time step, which supports the idea that gradients across time can be beneficial for end-to-end training. For larger $k$, or if we update the parameters of the (not frozen) resampler during end-to-end training, we see that performance degrades with higher $k$. The $k=5$ error rates of 9.9\% and 10.6\% suggest that end-to-end training is unstable because end-to-end training leads to only a slight improvement from the error rate of 12\% for individual training that we saw in Table~\ref{tab:main_table}. 

\begin{table}[ht!]
\caption{DPF test results with gradient stopping every $k$ steps}
\label{tab:mod}
\vspace{-0.3cm}
\centering{}%
\adjustbox{width=0.75\linewidth}{
    \begin{tabular}{ccccccc}
    \toprule
    && \multicolumn{2}{c}{Resampler frozen} & &  \multicolumn{2}{c}{Resampler not frozen} \\
    \cmidrule{3-4} \cmidrule{6-7}
    $k$ && Error rate & MSE & & Error rate & MSE \\
    \midrule
    1 && 4.7$\pm$0.2\% & 2.3$\pm$0.1 && {\bf1.2}$\pm$0.2\% & {\bf0.4}$\pm$0.1\\
    3 && {\bf 4.3}$\pm$0.2\% & {\bf2.1}$\pm$0.1 && 4.9$\pm$0.3\% & 1.8$\pm$0.1\\
    5 && 9.9$\pm$0.2\% & 2.8$\pm$0.1 && 10.6$\pm$0.6\% & 2.5$\pm$0.2\\
    \bottomrule
    \end{tabular}
}
\end{table}

\begin{table}[ht!]
\caption{DPF test results with gradient clipping and gradient stopping every $k$ steps}
\label{tab:gradient_clipping}
\vspace{-0.3cm}
\centering{}%
\adjustbox{width=0.75\linewidth}{
    \begin{tabular}{ccccc}
    \toprule
    $k$ && Resampler frozen & & Resampler not frozen \\
    \midrule
    1 && 4.7$\pm$0.2\% && {\bf1.1}$\pm$0.07\% \\
    2 && {\bf3.5}$\pm$0.1\% && 1.2$\pm$0.14\% \\
    3 && 3.7$\pm$0.2\% && 1.9$\pm$0.2\% \\
    4 && 3.9$\pm$0.1\% && 5.4$\pm$1.6\% \\
    5 && 4.4$\pm$0.1\% && 9.8$\pm$0.1\% \\
    \bottomrule
    \end{tabular}
}
\end{table}

\begin{figure}[ht!]
\includegraphics[width=1\columnwidth]{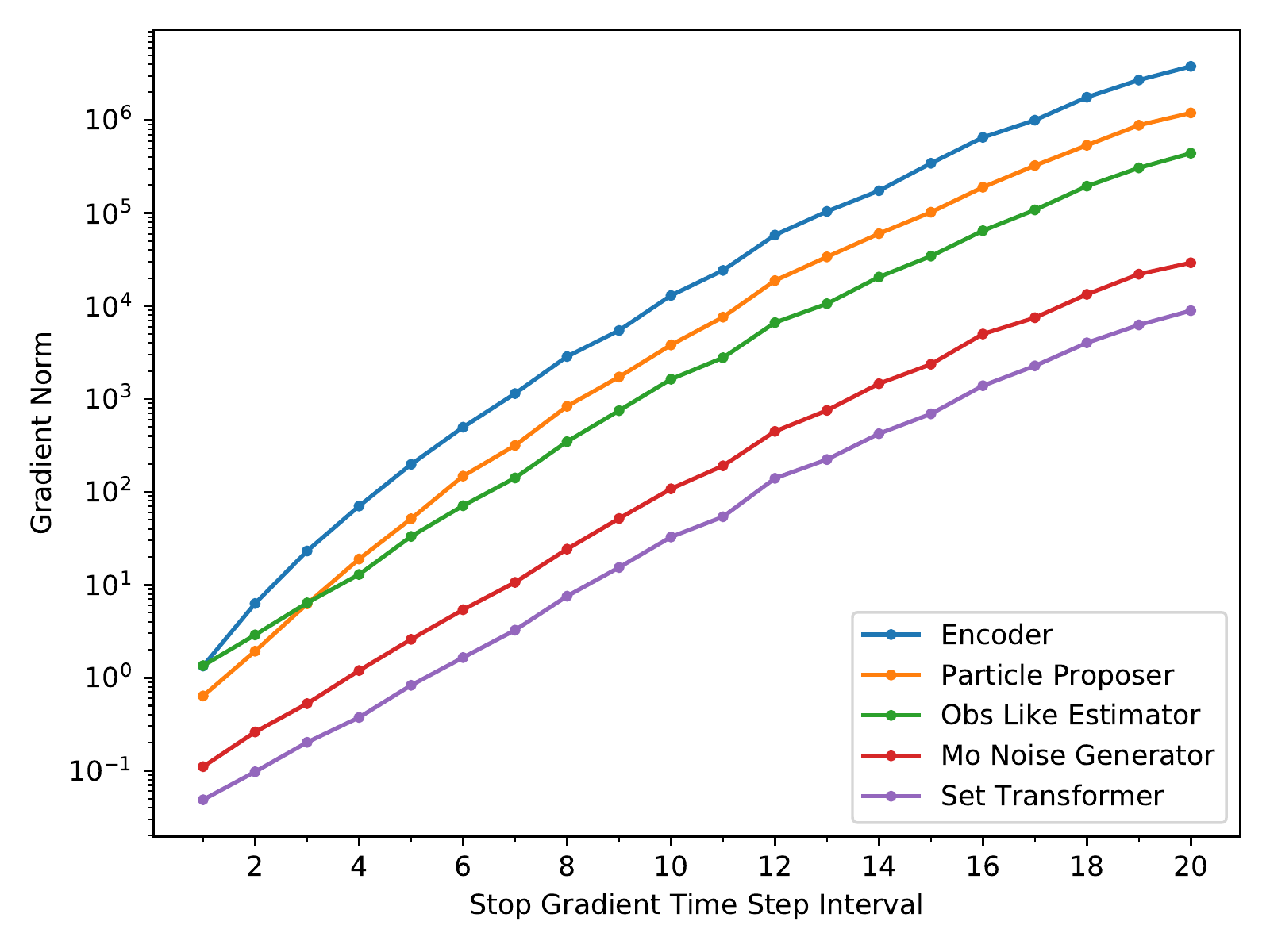}
\vspace{-0.8cm}
\caption{Median gradient norms for different DPF components as a function of $k$, the number of time steps after which BPTT is truncated.}
\label{fig:gradient_norms}
\vspace{-0.4cm}
\end{figure}

Suspecting exploding gradients to cause this instability, we analyzed how the median gradient norms for DPF components depend on $k$ (see Figure~\ref{fig:gradient_norms}). Note the logarithmic scale on the y-axis -- the gradient norms are almost exponential in $k$.

To solve this problem, we clipped the gradient norms at 10, which produced the results in Table~\ref{tab:gradient_clipping}. Comparing these results to Table~\ref{tab:mod} shows that gradient clipping generally improves performance, e.g. for $k = 3$ error rates reduce from 4.3\% to 3.7\% (resampler frozen) and from 4.9\% to 1.9\% (not frozen). With gradient clipping, BPTT for $k\in\{2,3,4\}$ provides a substantial improvement over $k=1$ if the resampler is frozen during end-to-end training. However, performance still deteriorates with larger $k$ and when the resampler is not frozen. The best results are still obtained by training the resampler end-to-end with $k=1$, i.e. not performing BPTT. While gradient clipping shows promise, more work is needed to properly train differentiable particle filters with particle transformers end-to-end across multiple time steps.

\begin{figure*}[ht!]
\centering
\subfloat[Input]{\includegraphics[width=0.33\linewidth]{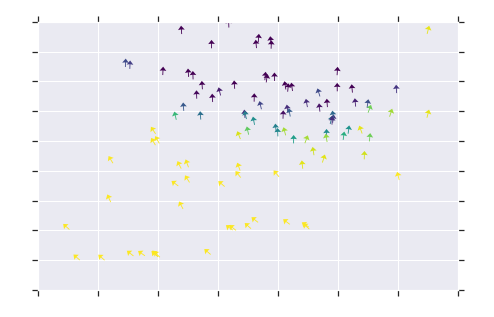}}
\subfloat[Systematic resampling]{\includegraphics[width=0.33\linewidth]{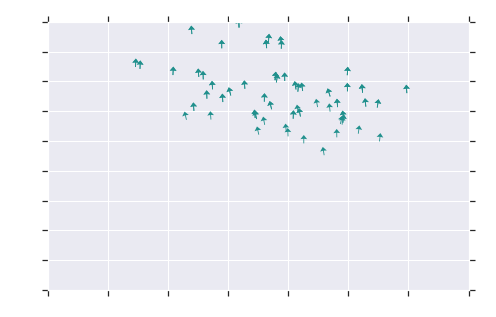}}
\subfloat[Particle transformer]{\includegraphics[width=0.33\linewidth]{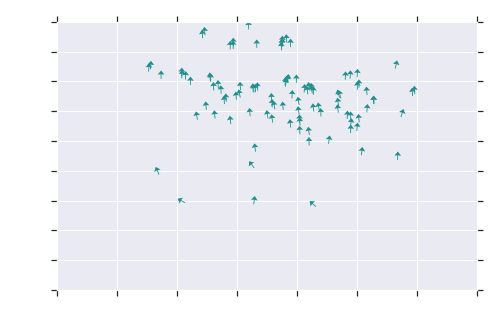}}
\\
\subfloat[Input]{\includegraphics[width=0.33\linewidth]{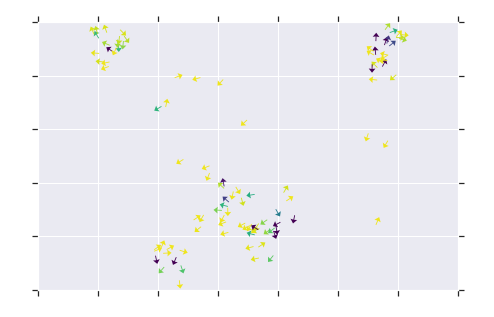}}
\subfloat[Systematic resampling]{\includegraphics[width=0.33\linewidth]{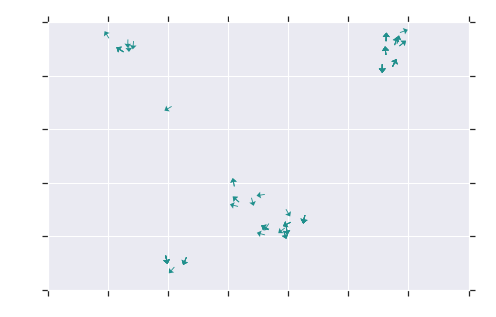}}
\subfloat[Particle transformer]{\includegraphics[width=0.33\linewidth]{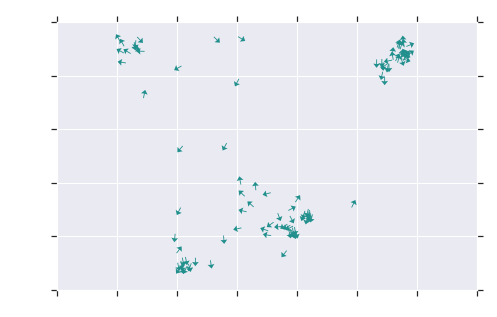}}
\\
\vspace{0.3cm}
\includegraphics[width=1\linewidth]{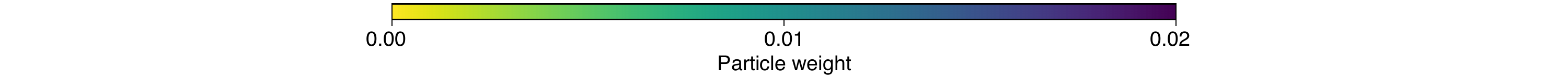}
\vspace{-0.6cm}
\caption{Qualitative results on two example particle sets in 2D position and angle. 
For a given input particle set (left), we show outputs of systematic resampling (middle) and the particle transformer (right). Particles are colored by their weights.}
\label{fig:dpf_comparing}
\end{figure*}

\subsection{Qualitative Results}
\label{app:qualitative_results}

\balance

We show quantitatively in Table~\ref{tab:main_table} that resampling with an end-to-end learned particle transformer leads to significantly better performance in the simulated localization task than using systematic resampling. To understand what the learned model does differently from the traditional resampler, we will now take a look at qualitative results.

Figure~\ref{fig:dpf_comparing} shows a side-by-side qualitative comparison of systematic resampling and resampling with the particle transformer for two examples. In the first example (top row), the input (a) consists of a set of highly weighted (purple) particles at the top and particles with weights close to zero (yellow) at the bottom. The results show that both the systematic resampler and the particle transformer sample most particles from the highly weighted region. But while the systematic resampler discards the regions of low weight entirely, the particle transformer produces a few samples there. In the second example (bottom row), the distribution of input particles (d) is multimodal. Compared to systematic resampling, we again see that the particle transformer places particles in regions without highly weighted inputs. Additionally, the particle transformer produces a more diverse set of particles near the modes of the distribution. 

It is plausible that generating additional particles in low weight regions and producing a more diverse particle set may increase robustness to modelling errors. As the motion and measurement models are learned from data, particles can, for example, get incorrectly weighted very low if the current combination of observation and state has not been seen during training. Particle transformers might improve performance by providing more robustness to these problems than traditional resampling methods. 

\begin{figure*}
\subfloat[Systematic resampling]{%
\includegraphics[width=1\linewidth]{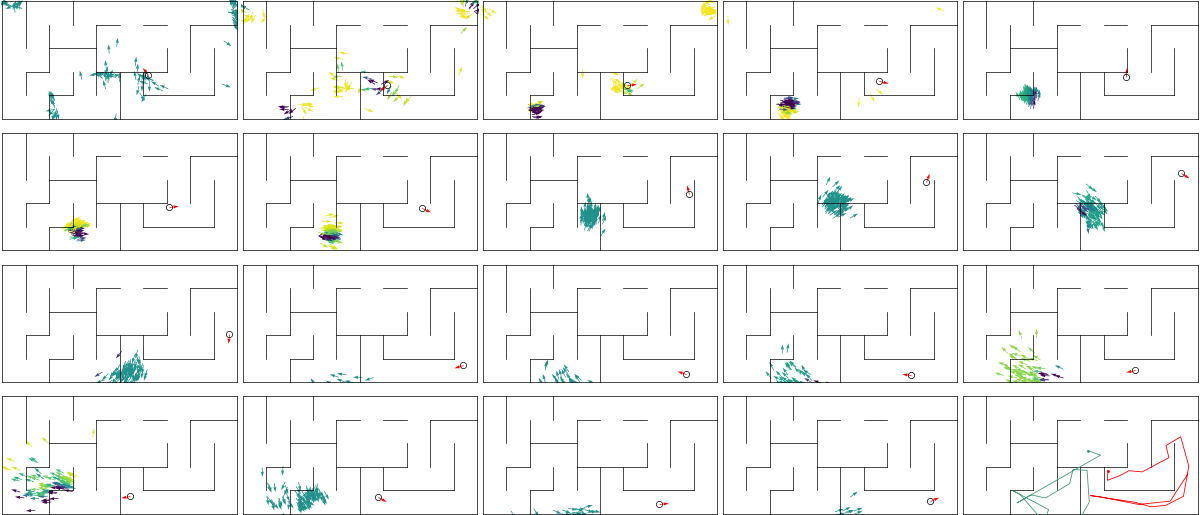}
}
\\
\subfloat[Particle transformer]{%
\includegraphics[width=1\linewidth]{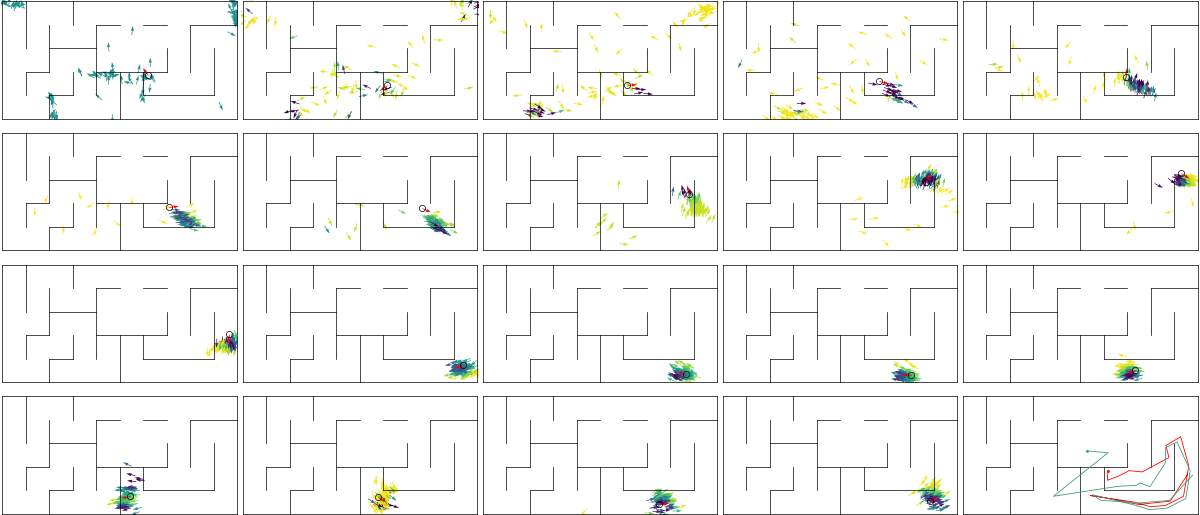}
}
\\
\begin{minipage}[t]{1\linewidth}
\vspace{0.2cm}
\includegraphics[width=1\linewidth]{images/colorbar}%
\end{minipage}
\caption{Visualization of the particle filter for robot localization (top-down view) on the same data with two different resampling methods. There are 19 time steps in total, 5 per row, with the last box summarizing the trajectories of the particle filter estimate (green) and the true robot pose (red). The true robot position and angle at each time step are shown by the black circle and red arrow. The particles colored by their weight represent the belief over where the robot is.}
\label{fig:dpf_visualization}
\end{figure*}

Figure~\ref{fig:dpf_visualization} shows one representative example of how increased robustness can have a major effect for the multi-step filtering process. We show particle filtering on the same sequence of observations and actions, once with systematic resampling (a), and once with the particle transformer (b). Both filtering runs start with a similar initial set of particles in step one (top left maze) and estimate a multi-modal posterior with roughly three modes in steps two and three. At step three, the more diverse set of particles in (b) allowed some particles near the true pose to receive high weights, while the weights in (a) are very focused on the wrong hypothesis (which from the local perspective of the robot looks indistinguishable). At step four, the particle filter with the traditional resampler (a) places almost no particles near the true pose and subsequently is not able to locate the robot. The particle filter with the learned resampler (b) distributes most particles across both valid hypotheses at step four, which later allows it to pick the correct one and track that hypothesis successfully.

%%%%%%%%%%%%%%%%%%%%%%%%%%%%%%%%%%%%%%%%%%%%%%%%%%%%%%%%%%%%%%%%%%%%%%%%%%%%%%%%
\pagebreak
\section*{Acknowledgments}
We would like to thank Anelia Angelova and Vincent Vanhoucke for their helpful comments on this work and Chad Richards for editing and proofreading the manuscript.
\balance
\bibliographystyle{plain}
\bibliography{references}

%%%%%%%%%%%%%%%%%%%%%%%%%%%%%%%%%%%%%%%%%%%%%%%%%%%%%%%%%%%%%%%%%%%%%%%%%%%%%%%%
\end{document}